\pdfoutput=1

\documentclass[11pt]{article}
\usepackage{graphicx}
\usepackage[final]{acl}
\usepackage{times}
\usepackage{latexsym}

\usepackage[T1]{fontenc}

\usepackage[utf8]{inputenc}

\usepackage{microtype}

\usepackage{inconsolata}
\usepackage[utf8]{inputenc}
\usepackage{listings}
\usepackage{pdfpages}

\usepackage{amsmath}
\usepackage{amsfonts}
\usepackage{geometry}
\usepackage{booktabs}
\usepackage{multirow}
\usepackage{ulem}
\usepackage{tcolorbox}
\tcbuselibrary{skins, breakable}
\newtcolorbox[auto counter, number within=section]{promptbox}[2][]{%
    colframe=blue!75!black,
    colback=blue!10,
    coltitle=white,
    fonttitle=\small\bfseries,
    title=Prompt Template~\thetcbcounter: #2,
    breakable, 
    enhanced,
    fontupper=\ttfamily,
    #1 
}

\title{\textsc{LLM-Hanabi}: Evaluating Multi-Agent Gameplays with Theory-of-Mind and Rationale Inference in Imperfect Information Collaboration Game
}

\author{Fangzhou Liang\thanks{~~Equal Contribution}, Tianshi Zheng\footnotemark[1], Chunkit Chan, Yauwai Yim, Yangqiu Song \\
  Department of Computer Science and Engineering, HKUST, Hong Kong SAR, China\\
  \texttt{fliangae@connect.ust.hk}\\
}

\begin{document}
\maketitle
\begin{abstract}
Effective multi-agent collaboration requires agents to infer the rationale behind others' actions, a capability rooted in Theory-of-Mind (ToM). While recent Large Language Models (LLMs) excel at logical inference, their ability to infer rationale in dynamic, collaborative settings remains under-explored. This study introduces \textsc{LLM-Hanabi}\footnote{\href{https://github.com/HKUST-KnowComp/LLM-Hanabi}{https://github.com/HKUST-KnowComp/LLM-Hanabi}}, a novel benchmark that uses the cooperative game Hanabi to evaluate the rationale inference and ToM of LLMs. Our framework features an automated evaluation system that measures both game performance and ToM proficiency. Across a range of models, we find a significant positive correlation between ToM and in-game success. Notably, first-order ToM (interpreting others' intent) correlates more strongly with performance than second-order ToM (predicting others' interpretations). These findings highlight that for effective AI collaboration, the ability to accurately interpret a partner's rationale is more critical than higher-order reasoning. We conclude that prioritizing first-order ToM is a promising direction for enhancing the collaborative capabilities of future models.
\end{abstract}

\section{Introduction}
The reasoning and planning capabilities of Large Language Models (LLMs) have advanced rapidly, leading to strong logical inference abilities. The next frontier is applying these capabilities to multi-agent collaboration, which demands not just logic, but also rationale inference—the ability to deduce the reasons behind another agent's actions. This process is fundamentally enabled by Theory-of-Mind (ToM), the cognitive capacity to attribute mental states like beliefs and intentions to others \cite{premack1978does}.

However, existing benchmarks for evaluating ToM in LLMs are often limited. Many rely on static, text-based tasks like story question-answering \cite{zhou2023far, he2023hitombenchmarkevaluatinghigherorder, chen2024tombenchbenchmarkingtheorymind}, which fail to capture the dynamic and uncertain nature of real-world collaboration. Other multi-agent frameworks may lack the scalability needed for large-scale, automated evaluation \cite{xu2024opentomcomprehensivebenchmarkevaluating}. This creates a gap in our understanding of how LLMs reason and cooperate in interactive settings with imperfect information.

To bridge this gap, we introduce \textsc{LLM-Hanabi}, a benchmark designed to assess rationale inference and ToM in a dynamic, collaborative environment. We use the cooperative card game Hanabi, where players have incomplete information and must rely on interpreting sparse linguistic hints to succeed. The game's core mechanics make it an ideal testbed for evaluating collaborative reasoning under uncertainty. Our framework translates game states into natural language, allowing LLM-driven agents to interact with the game environment and each other, with an automated system for scalable evaluation.

Our contributions are threefold:
\vspace{-0.1cm}
\begin{enumerate}
\item We develop \textbf{\textsc{LLM-Hanabi}}, an automated framework for evaluating rationale inference and ToM in a dynamic, multi-agent setting.
\vspace{-0.1cm}
\item We benchmark a diverse set of LLMs, providing a comprehensive analysis of their collaborative performance.
\vspace{-0.1cm}

\item We demonstrate a strong positive correlation between ToM and game success, and reveal that \textbf{first-order ToM is a more significant predictor of performance than second-order ToM}. 
\end{enumerate}
\begin{figure*}[h!]
    \centering
    \includegraphics[width=\textwidth]{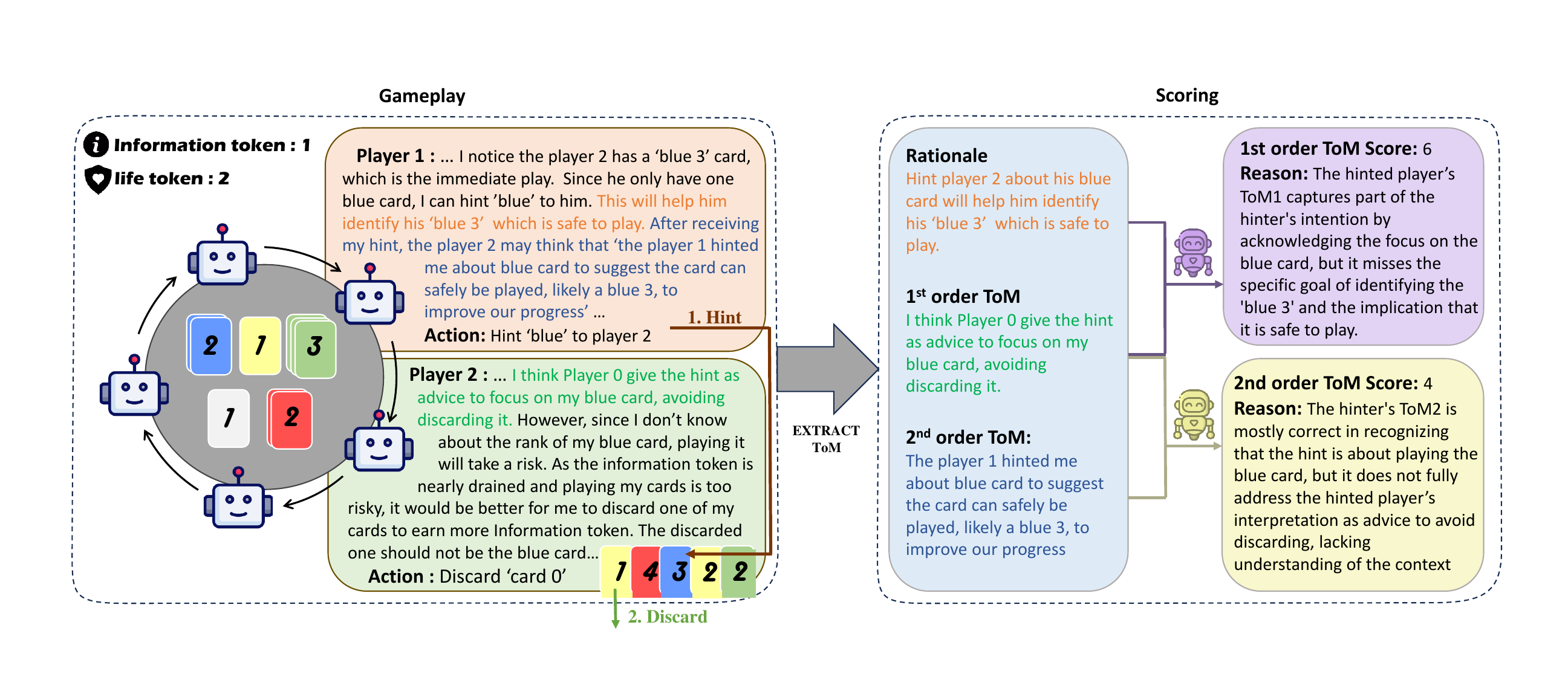} 
    \caption{Overview of the gameplay and evaluation process of \textsc{LLM-Hanabi}.}
    \label{fig}
\vspace{-0.2cm}

\end{figure*}

\section{Related Work}
\subsection{LLMs in Multi-Agent Gameplays} Imperfect information games are a powerful tool for testing the reasoning abilities of LLMs in scenarios that mimic real-world complexity \cite{yim2024evaluating}. Recent work shows that LLMs can perform effective multi-step reasoning, sometimes surpassing traditional reinforcement learning methods in settings like two-player games \cite{guo2024suspicionagentplayingimperfectinformation} and open-world survival challenges \cite{humanbehaviour}. Coordination benchmarks have also been developed to study agent interactions in various simulated environments \cite{carroll2020utilitylearninghumanshumanai, wang2020cooksbayesianinferencecoordinating}.

\subsection{Theory-of-Mind} Theory-of-Mind, the ability to attribute mental states to others \cite{frith2005theory}, is essential for navigating imperfect information games. The ToM capabilities of LLMs have recently been evaluated using various textual tasks, primarily focusing on question-answering scenarios \cite{zhou2023far, kim2023fantom, chen2024tombenchbenchmarkingtheorymind, xu2024opentomcomprehensivebenchmarkevaluating,chan2024negotiationtombenchmarkstresstestingmachine,chan2025xtomexploringmultilingualtheory}. These studies, including explorations into higher-order ToM \cite{he2023hitombenchmarkevaluatinghigherorder}, have deepened our understanding of how LLMs process complex social and interactive information.

\section{\textsc{LLM-Hanabi} Benchmark}
\subsection{Hanabi Game Overview}

Hanabi is a cooperative card game for 2-5 players. The goal is to build five color-coded stacks of cards in ascending order (1 to 5). The central challenge is that players cannot see their own cards, creating a game of imperfect information. On their turn, a player can perform one of three actions: (1) play a card, (2) discard a card to replenish a hint token, or (3) give a hint to a teammate about the color or number of their cards. The game begins with eight hint tokens and three life tokens. A life token is lost if a card is played incorrectly. The game ends when all stacks are complete, the deck runs out (triggering a final round), or all life tokens are lost. The final score is the sum of the highest card values in each of the five stacks. Success hinges on effective communication and inferring intent from limited hints.

\begin{table*}[t]
\centering
\small
\begin{tabular}{llcccccc}
\toprule
\multirow{2}{*}{\centering \textbf{Category}} & \multirow{2}{*}{\textbf{Models}} 
& \multicolumn{3}{c}{\textbf{Game Performance}} 
& \multicolumn{3}{c}{\textbf{ToM Performance}} \\
\cmidrule(lr){3-5} \cmidrule(lr){6-8}
& & \textbf{Score (Std.)} & \textbf{Min/Max} & \textbf{\#Rounds}
  & \textbf{1st-order} & \textbf{2nd-order} & \textbf{Average} \\
\midrule
\multirow{9}{*}{\centering LLM (CoT)} & Llama-3.1-8B 
  & 3.47 (4.42) & 0-16 & 10 & 61.98 & 20.71 & 41.34 \\
 & Llama-3.1-70B 
  & 4.13 (4.51) & 0-20&12 &72.45 &53.10 & 62.78\\
 & Llama-3.1-405B 
  & 5.84 (5.67) & 0-24 & 13 & 77.63 & 57.66 & 67.64 \\
 & Llama-3.3-70B 
  & 7.20 (4.97) & 0-20 & 17 & 81.74 & \underline{69.79} & 75.76 \\
 & Llama-4-Scout 
  & 5.12 (3.96) & 0-16 & 26 & 77.41 & 57.03 & 67.22 \\
 & Llama-4-Maverick 
  & 20.08 (6.49) & 8-36 & 19 & 85.21 & 58.83 & 72.02 \\
 & gpt-4.1-nano
  & 5.36 (3.84) & 0-12 & 18 & 66.55 & 52.03 & 59.29 \\
 & gpt-4.1-mini
  & 25.60 (6.67) & 8-36 & 22 & \underline{86.70} & 65.27 & 75.99 \\
 & gpt-4.1
  & \underline{28.56} (6.10) & 12-40 & 23 & 83.81 & \textbf{72.46} & \textbf{78.14} \\

\midrule
\multirow{5}{*}{\centering LRM (Long-CoT)} 
 & Qwen3-8b
  & 22.88 (6.10) & 8-36 & 24 & 82.29 & 60.73 & 71.51 \\
 & Qwen3-30b
  & 19.60 (5.38) & 4-28 & 26 & 72.60 & 54.47 & 63.54 \\
 & QwQ-32B 
  & 28.27 (4.32) & 20-36 & 25 & 84.48  & 59.69 & 72.09 \\
& Gemini-2.5-Flash 
  & 25.04 (4.97) & 16-36 & 26 & 86.67 & 67.13 & 76.90 \\
 & Deepseek-R1 
  & \textbf{30.00 (3.45)} & 24-36 & 26 & \textbf{87.05} & 68.31 & \underline{77.68} \\

\bottomrule
\end{tabular}
\caption{Performance Results of LLM and LRM Agents in \textsc{LLM-Hanabi} (100-point scale). Bold indicates best, underline indicates second-best in each column.}
\label{tab:results}
\vspace{-0.2cm}
\end{table*}

\subsection{\textsc{LLM-Hanabi} Design Rationale}
Hanabi is an ideal environment for studying rationale inference. Its partial observability forces agents to infer information from sparse linguistic cues, while the hint-giving mechanism provides a natural way to probe their reasoning. Unlike adversarial games like Poker or Werewolf, which involve deception and complex strategies, Hanabi's purely cooperative nature isolates the challenge of collaborative reasoning. This allows for a more direct evaluation of an agent's ability to infer and act on a partner's intentions. \textsc{LLM-Hanabi} is designed to leverage these features, creating a controlled testbed for assessing ToM and rationale inference.

\begin{figure}[t]
    \centering
    \includegraphics[width=1\linewidth]{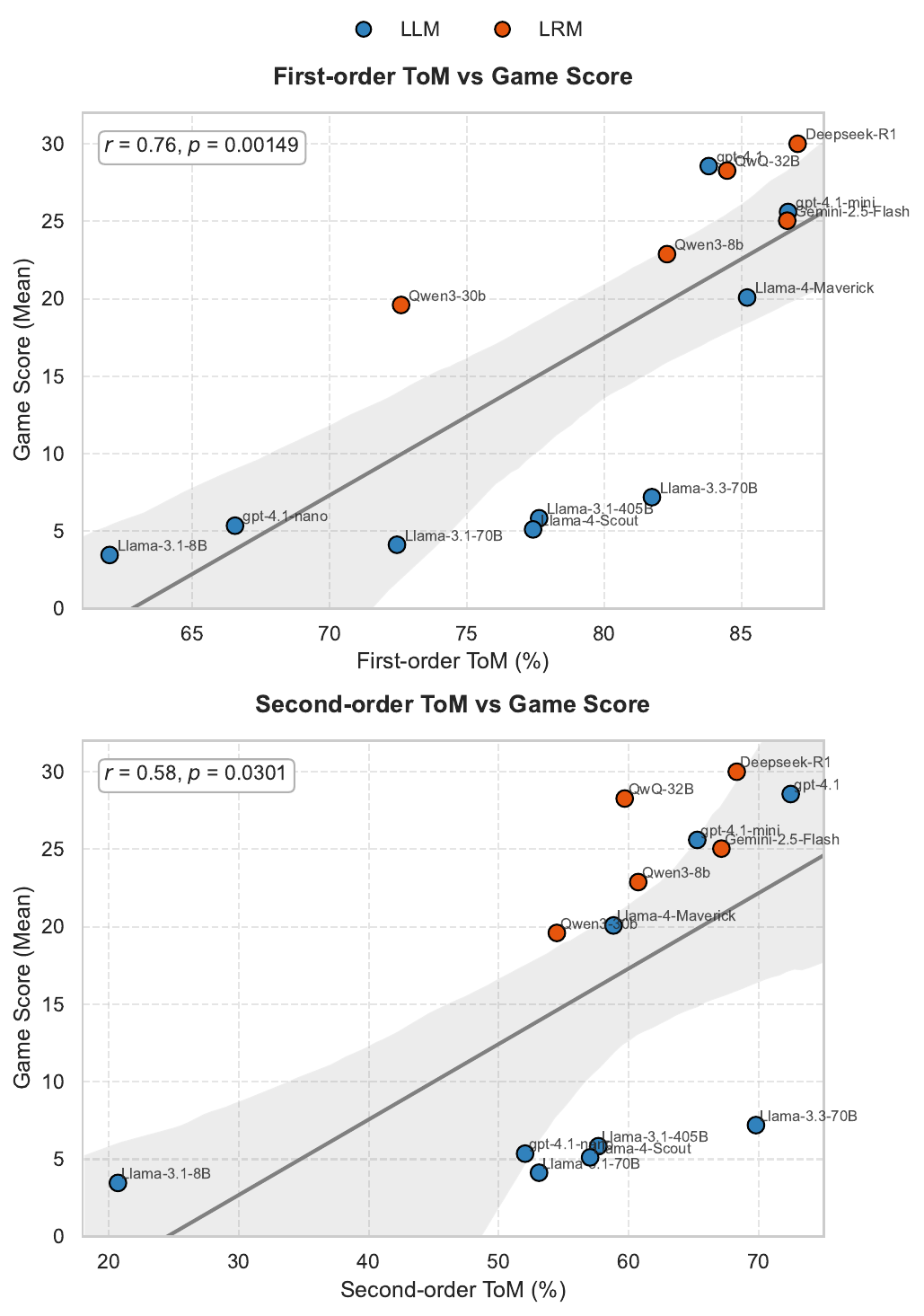}
    \caption{Correlation analysis between game score and theory-of-mind performance.}
    \label{fig:regression}
\vspace{-0.3cm}
\end{figure}

\subsection{ToM Evaluation System}

As illustrated in Figure \ref{fig}, our benchmark evaluates ToM through a two-phase process: reasoning extraction during gameplay and post-game scoring via an LLM-as-a-judge. 

\paragraph{Reasoning Information Extraction} During each hint action, agents are prompted to generate structured statements capturing their reasoning process:
\begin{itemize}
\item \textbf{Rationale:} The hinter's justification for giving a specific hint. This serves as the ground truth for the hint's intent.
\item \textbf{First-Order ToM:} The hinted player's interpretation of the hinter’s intent.
\item \textbf{Second-Order ToM:} The hinter’s prediction of how the hinted player will interpret the hint.
\end{itemize}

\paragraph{Post-Game Scoring} After each game, an LLM-as-a-judge evaluates the extracted statements to produce ToM scores on a 0-10 scale:
\begin{itemize}
\item \textbf{First-Order ToM Score:} Measures the alignment between the hinter's \textit{Rationale} and the recipient's \textit{First-Order ToM} interpretation. This reflects the recipient's inferential accuracy.
\item \textbf{Second-Order ToM Score:} Measures the alignment between the recipient's \textit{First-Order ToM} and the hinter's \textit{Second-Order ToM} prediction. This reflects the hinter's predictive accuracy.
\end{itemize}
This system provides a scalable and quantitative method for assessing interactive ToM and rationale inference.

\section{Evaluation}

This section evaluates a range of LLMs and LRMs using the \textsc{LLM-Hanabi} benchmark. We analyze their performance based on Game Scores and ToM Scores to understand the relationship between their reasoning abilities and collaborative success.

\subsection{Models}

We evaluated a diverse set of LLMs and LRMs, with details in Appendix \ref{app:models}. All models were prompted (details in Appendix \ref{app:prompt}) to use Chain-of-Thought (CoT) reasoning to inform their decisions and generate the required ToM statements.

\subsection{Evaluation Metrics}

The \textsc{LLM-Hanabi} benchmark employs a structured evaluation framework with two primary metrics to assess logical inference capabilities. The following details experimental configurations and metrics.

\paragraph{Game Settings} All experiments were conducted in a 5-player configuration to maximize interaction complexity. Each model type played 30 to 50 games to ensure statistical significance. The game environment followed standard Hanabi rules, with 8 information tokens and 3 life tokens.
\paragraph{Recorded Metrics} The following metrics were collected during each game to provide a comprehensive evaluation of agent performance:
\begin{itemize}
\item\textbf{Game Score:} This score is the aggregate of the highest card ranks successfully played across the five firework stacks, with a maximum possible score of 25. It serves as the primary measure of the agents' collective ability to achieve the game's cooperative objective through strategic play.
\item\textbf{ToM Score:} This metric is the arithmetic mean of the first-order and second-order scores generated across all hint interactions within a game. It quantifies the agents’ proficiency in rationale inference and reasoning about others' mental states.
\end{itemize}
Following the experiments, a correlation analysis was performed between Game Scores and ToM Scores to elucidate the relationship between ToM capabilities and overall cooperative success.

\subsection{Results and Analyses}
Our evaluation of various LLMs and LRMs reveals significant performance differences in both gameplay and ToM capabilities. The results, detailed in Table~\ref{tab:results}, consistently show a strong correlation between an agent's ToM proficiency and its success in cooperative gameplay.

\paragraph{General Game Performance} LRMs demonstrated markedly superior gameplay, achieving higher average scores than most LLMs, which indicates a greater capacity for effective cooperation within the Hanabi environment. Among LRM agents, Deepseek-R1 achieved the highest average score (30.00), with QwQ-32B also delivering an exceptional performance (28.27). While performance in the LLM category was more varied, gpt-4.1 (28.56) and gpt-4.1-mini (25.60) were notable outliers, performing at a level comparable to the top-performing LRMs.

\paragraph{ToM Performance} Consistent with the game score results, LRMs generally outperformed LLMs in our ToM assessments, reflecting a more robust ability to reason about the mental states of other agents. The top-performing models in this category were gpt-4.1 from the LLM group (average ToM score of 78.14) and Deepseek-R1 from the LRM group (77.68).

\paragraph{First-order vs. Second-order ToM} Across all models, a distinct pattern emerged: first-order ToM scores were consistently and significantly higher than second-order scores. This disparity suggests that while models are generally proficient at interpreting the direct intent of a given hint (first-order), they are far less capable of accurately predicting how another agent will, in turn, interpret that hint (second-order).

\paragraph{Correlation between ToM and Game Performance} As illustrated in Figure \ref{fig:regression}, our analysis confirms a strong, positive correlation between ToM proficiency and game performance. Higher ToM scores are consistently associated with better collaborative outcomes. Most notably, we found that first-order ToM is more significantly correlated with game success (r=0.76) than second-order ToM (r=0.58). This critical finding suggests that for successful collaboration, the recipient's accurate inference of a hint's rationale is more impactful than the hinter's prediction of the recipient's interpretation.

\section{Conclusion}
This study introduced \textsc{LLM-Hanabi}, a benchmark for evaluating rationale inference and Theory-of-Mind in collaborative agents. Our results yield two key insights: (1) Large reasoning models significantly outperform standard LLMs in both gameplay and ToM, demonstrating the value of specialized reasoning architectures. (2) First-order ToM (rationale inference) is a far stronger predictor of success than second-order ToM. These findings suggest a clear path forward: developing effective collaborative AI hinges more on an agent's ability to accurately interpret its partners than on modeling higher-order beliefs. Future studies could leverage this benchmark to develop and evaluate methodologies for enhancing these specific inferential skills.

\section*{Limitations}
While \textsc{LLM-Hanabi} offers a robust and scalable framework for evaluating rationale inference and Theory-of-Mind capabilities in language models, several limitations remain. First, the benchmark is restricted to the Hanabi game environment, which, though ideal for controlled ToM evaluation, may not capture the full diversity of collaborative scenarios encountered in real-world multi-agent systems. Second, the evaluation primarily focuses on language-based reasoning and may not fully reflect the models’ performance in multimodal or non-linguistic contexts. Third, the reliance on an LLM-as-a-judge for ToM scoring introduces potential bias and subjectivity, as the assessment quality depends on the judge model’s own reasoning abilities. Finally, our experiments are conducted under fixed hyperparameters (e.g., temperature) and specific prompt formats, which may affect generalizability across different settings and downstream applications. Future work should explore broader environments, alternative evaluation modalities, and human-in-the-loop assessments to further validate and extend the findings.

\section*{Ethics Statement}
This research is conducted with the intention of advancing understanding of collaborative reasoning and Theory-of-Mind in large language models, with a focus on safe and beneficial AI development. All experiments are performed in simulated environments without involving human subjects or personally identifiable information. Nevertheless, we acknowledge that improvements in multi-agent reasoning and ToM capabilities could be misused in adversarial or manipulative contexts, such as deceptive agent design or privacy-invasive applications. We encourage the responsible deployment of such technologies and recommend that future work incorporate explicit safeguards and alignment strategies. Researchers and practitioners using the \textsc{LLM-Hanabi} benchmark should remain vigilant about the ethical implications of enhanced AI collaboration and inference, ensuring that developments contribute positively to society and respect user autonomy and privacy.

\bibliography{main}
\bibliographystyle{acl_natbib}

\appendix

\newpage
\section{Model Details}
\label{app:models}
In our experiments, we tested 14 modern LLM / LRMs with their detailed information as follows:
\begin{itemize}
    \item \textbf{Llama-3.1-8b / Llama-3.1-70b / Llama-3.1-405b} \cite{meta2024llama31} is an open-source dense LLM series, developed with a pre-training corpus of 15 trillion tokens, incorporating DPO during alignment to enhance performance.
    \item \textbf{Llama-3.3-70b} \cite{meta2024llama33} is an open-source dense LLM, refined from the Llama series with an expanded training dataset, optimized for improved reasoning capabilities.
    \item \textbf{Llama-4-Scout / Llama-4-Maverick} \cite{meta2024llama4} is an open-source dense LLM series, designed with advanced scaling techniques and trained on a diverse corpus to support complex reasoning tasks.
    \item \textbf{Deepseek-R1-671b} \cite{deepseekai2025deepseekr1incentivizingreasoningcapability} is a leading open-source LRM, trained using reinforcement learning with a rule-based reward system to boost logical inference.
    \item \textbf{QwQ-32b} \cite{qwen2025qwq32b} is an open-source LRM from Alibaba, engineered with a focused training approach to balance efficiency and performance in multi-agent scenarios.
    \item \textbf{Qwen3-8b / Qwen3-30b} \cite{yang2025qwen3technicalreport} is an open-source Mixture-of-Experts (MoE) LRM series, built with a pre-training corpus exceeding 18 trillion tokens and fine-tuned on extensive datasets.
    \item \textbf{GPT-4.1-nano / GPT-4.1-mini / GPT-4.1} \cite{openai2024gpt4technicalreport} is a proprietary LLM series by OpenAI, representing their latest pre-reasoning models with undisclosed training details.
    \item \textbf{Gemini-2.5-flash} \cite{comanici2025gemini25pushingfrontier} is a proprietary Mixture-of-Experts (MoE) LRM, featuring enhanced multimodal capabilities and a long context window of 1 million tokens.
\end{itemize}
The temperature for all LLM/LRMs is set to 0.7 in our main experiments.

\begin{table}[h]
\centering
\scriptsize
\begin{tabular}{lll}
\toprule
\textbf{Series} & \textbf{Creator} & \textbf{\# Parameters} \\
\midrule
\multicolumn{3}{c}{\emph{Open-source LLMs}} \\
\addlinespace[0.5em]
Llama-3.1 \cite{meta2024llama31} & Meta & 8B, 70B, 405B \\
Llama-3.3 \cite{meta2024llama33} & Meta & 70B \\
Llama-4-Scout \cite{meta2024llama4} & Meta & 109B \\
Llama-4-Maverick \cite{meta2024llama4} & Meta & 400B \\
Deepseek-R1 \cite{deepseekai2025deepseekr1incentivizingreasoningcapability} & Deepseek & 671B \\
QwQ \cite{qwen2025qwq32b} & Alibaba & 32B \\
Qwen3 \cite{yang2025qwen3technicalreport} & Alibaba & 8B, 30B \\
\addlinespace[0.5em]
\multicolumn{3}{c}{\emph{Proprietary LLMs}} \\
\addlinespace[0.5em]
GPT-4.1-nano \cite{openai2024gpt4technicalreport} & OpenAI & - \\
GPT-4.1-mini \cite{openai2024gpt4technicalreport} & OpenAI & - \\
GPT-4.1 \cite{openai2024gpt4technicalreport} & OpenAI & - \\
Gemini-2.5-flash \cite{comanici2025gemini25pushingfrontier} & Google & - \\
\bottomrule
\end{tabular}
\caption{LLMs evaluated in our experiments.}
\label{tab:models}
\end{table}

\newpage
\section{Related Works on Rationale Inference}
The inference of decision rationales is often associated with abductive reasoning, a process wherein an agent must generate an explanatory hypothesis that logically entails a given observation~\cite{bai2024advancingabductivereasoningknowledge,gao2025controllablelogicalhypothesisgeneration,zheng2025logidynamicsunravelingdynamicsinductive,zheng2025enhancingtransformersgeneralizablefirstorder}. Furthermore, the capability for abductive reasoning can be developed through agentic post-training, wherein agents inductively learn to connect decision rationales with their corresponding outcomes~\cite{li2025patternsprinciplesfragilityinductive,fan2025legalruleinductiongeneralizable,zheng2025cursecotlimitationschainofthought}. In turn, these developed capabilities enhance an agent's overall performance and autonomy when operating in complex, real-world environments~\cite{xu2025multiagentreasoningsystemscollaborative,zheng2025automationautonomysurveylarge}.
\newpage
\section{Prompt Details}
\label{app:prompt}

\begin{promptbox}[colback=black!10, colframe=white!50!black, title=Hanabi Game Rules]{}
\scriptsize
\begin{lstlisting}
# Core Mechanics
1. BASIC COMPONENTS:
   - Cards: 
     - Each player will get 3-5 cards in their hands. Each time when you play or discard your card, a new card will be drawn for you to keep the amount
     - Each card has its Color (red/yellow/green/blue/white) and Rank (1-5)
     - 50 cards in total (For each color, there are 3 rank-1 cards, 2 rank-2 cards, 2 rank-3 cards, 2 rank-4 cards and 1 rank-5 card in total)
   - Fireworks: 
     - Contain five color stack(red, yellow, green, blue, white)
     - The card should be contributed to the stack strictly in ascending numerical order.
   - Game History:
     - It records some of previous actions taken by different players
     - It serves as an important reference for making your current decision
   - Information Tokens: Used to give hints about the cards in their hands
   - Life Tokens: Lost 1 token when incorrectly playing a card (in wrong order)
2. SETUP:
   - Shuffle the cards and deal a specific number of cards to each player (2 cards for 2 players, 3 for 3 players, 4 for 4 players, 5 for 5 players). Remaining cards form draw pile 
3. PLAYER ACTIONS:
   Players take turns in order. On your turn, you can perform one of the three actions:
   - PLAY:
     - Play a card in your hand means trying to contribute it to the firework stacks
     - If played correctly, it contributes to the fireworks stacks
     - If wrong, life tokens will lose by 1. The card will then be discarded 
   - DISCARD:
     - Discard a card to the trash pile to regain 1 information token
   - HINT:
     - Hint another player about their cards' color or rank (you cannot hint yourself)
     - Spend an information token each time you hint.
     - Examples:
       - "Hint White": All white cards in the specified teammate's hand will be revealed to him
       - "Hint 1": All 1-rank cards in the specified teammate's hand will be revealed to him
       
# Game Over Conditions
    - Build all color stacks to 5 / No more deck / Life tokens reach 0 
    - Total score = sum of the largest rank of five firework stacks
\end{lstlisting}
\end{promptbox}

\newpage
\begin{promptbox}[colback=black!10, colframe=white!50!black, title=Game State Example]{}
\scriptsize

\begin{lstlisting}
"round": {
    "description": "An integer keeping track of the current round in the game. Start from 1.",
    "content": 1
},
"fireworks": {
    "description": "A dictionary showing the state of firework stacks, the numbers show the current highest rank of the corresponding color.",
    "content": {
        "red": 0,
        "blue": 0,
        "green": 0,
        "white": 0,
        "yellow": 0
    }
},
"knowledge": {
    "description": "A list showing your own view of the cards in all players' hands. Each card is represented by a tuple i.e. (COLOR, RANK). The symbol '?' means unknown color or unknown rank. The card index from 0 to 4, with the leftmost having an index of 0.",
    "content": {
        "The cards in your own hands": "[('?', '?'), ('?', '?'), ('?', '?'), ('?', '?'), ('?', '?')]",
        "The cards in Player_id 1's hands from your view": "[('yellow', 2), ('red', 1), ('white', 4), ('red', 5), ('white', 3)]",
        "The cards in Player_id 2's hands from your view": "[('yellow', 4), ('red', 1), ('green', 2), ('yellow', 1), ('red', 3)]",
        "The cards in Player_id 3's hands from your view": "[('blue', 3), ('white', 5), ('blue', 1), ('blue', 5), ('blue', 4)]",
        "The cards in Player_id 4's hands from your view": "[('red', 2), ('red', 3), ('green', 5), ('blue', 1), ('yellow', 1)]"
    }
},
"game_history": {
    "description": "A list of dictionaries showing the previous 10 actions of different players.",
    "content": [
        "In round 1, the player 0 HINT the player 2 about his card(s) of RANK 1"
    ]
},
"information_tokens": {
    "description": "An integer showing the current number of information tokens",
    "content": 8
},
"life_tokens": {
    "description": "An integer showing the current number of life tokens",
    "content": 3
}
\end{lstlisting}

\end{promptbox}
\newpage

\begin{promptbox}[colback=black!10, colframe=white!50!black, title=General Prompt]{}
\scriptsize
\begin{lstlisting}
Here is the Hanabi Game Rule: {Hanabi_Game_Rules}

You are playing hanabi game collectively 
with other 4 players. Your player_id is
0(count from 0).

The following dictionary shows the current 
state: {Game_State}

Now is your turn. If you want to give a 
hint to others, you should show your 2nd 
order Theory-of-Mind(ToM), which is your 
understanding of the hinted player's 
throught towards your hint.

Before your final reply, give me your chain
of thought. You should also show your 1st 
order ToM and 2nd order ToM if we tell you 
to do that.

Your reply should end with a separate json
**only** including the following keys, 
without any comments or irrelative 
sentences:
            {{
  "action_type": choose from 'PLAY'/ 'DISCARD'/'HINT_COLOR'/'HINT_RANK',
  "card_index": an integer from 0 to 4,  showing the card index to 'PLAY' or 'DISCARD' in your hands. For other actions, put "N/A" here
  "hint_player": an integer from 0 to {num_players-1}, showing the player you want to 'HINT'. For other actions, put "N/A" here
  "hint_color": a word chose from (Red, Blue, Green, Yellow, White),  showing the color you want to 'HINT'.  For other actions, put "N/A" here
  "hint_rank": an integer from 0 to 4, showing the rank you want to 'HINT'. For other action, put "N/A" here
  "rationale": a string, only applicable when you hint other. Show why you want to hint him. For other actions, put "N/A" here
  "1st_order_ToM": a string, only applicable when you are hinted by others. Show your understanding of why others hint you. For other actions, put "N/A" here
  "2nd_order_ToM": a string, only applicable when you hint other. Show your understanding of the hinted player's throught towards your hint . For other actions, put "N/A" here
            }}
Example:
           {{
  "action_type":"PLAY",
  "card_index":0,
  "hint_player":"N/A",
  "hint_color":"N/A",
  "hint_rank":"N/A",
  "rationale":"N/A",
  "1st_order_ToM":"N/A",
  "2nd_order_ToM":"N/A"
             ""
           }}
\end{lstlisting}
\end{promptbox}

\newpage
\begin{promptbox}[colback=black!10, colframe=white!50!black, title=Prompt for Hinted Player]{}
\scriptsize
\begin{lstlisting}
In addition to the General Prompt, the 
following is designed to elicit a
first-order Theory-of-Mind (ToM):

The following is the most recent hint you 
received:
    The player 3 hinted you about your 
    YELLOW card(s).
According to this/these hint(s), you should
show your 1st order Theory-of-Mind(ToM), 
which is your understanding of why he/they 
hint you. Then your decision should be 
based on these understanding.
\end{lstlisting}
\end{promptbox}

\end{document}